%% file: main.tex
\documentclass[10pt,twocolumn,letterpaper]{article}

\usepackage[pagenumbers]{cvpr} 

\input{preamble}

\definecolor{cvprblue}{rgb}{0.21,0.49,0.74}
\usepackage[pagebackref,breaklinks,colorlinks,citecolor=cvprblue]{hyperref}
\input{figs}
\input{tabs}

\usepackage{booktabs}

\def\paperID{214} 
\def\confName{3DV\xspace}
\def\confYear{2026\xspace}

\title{Broadening View Synthesis of Dynamic Scenes\\ from Constrained Monocular Videos}

\author{
Le Jiang \qquad Shaotong Zhu \qquad Yedi Luo \qquad Shayda Moezzi \qquad Sarah Ostadabbas\\
Northeastern University, Boston, MA\\
{\tt\small \{jiang.l, zhu.shaot, luo.ye, moezzi.s, s.ostadabbas\}@northeastern.edu}
}

\begin{document}
\maketitle
\input{sec/0_abstract}    
\input{sec/1_intro}
\input{sec/2_relatedwork}
\input{sec/3_method}
\input{sec/4_results}
\input{sec/5_conclusion}
{
    \small
    \bibliographystyle{ieeenat_fullname}
    \bibliography{main}
}
\input{sec/X_suppl}

\end{document}

%% file: preamble.tex
\usepackage[dvipsnames]{xcolor}


%% file: tabs.tex
\usepackage{amssymb} 
\usepackage{pifont}  
\usepackage{xcolor}  
 \usepackage{multirow} 
 \usepackage{float}

\newcommand{\cmark}{\textcolor{darkgreen}{\ding{51}}} 
\newcommand{\xmark}{\textcolor{red}{\ding{55}}}       

\definecolor{darkgreen}{rgb}{0.0, 0.5, 0.0}

\newcommand{\syndmtest}{%
\begin{table*}[!t]
  \centering
  \scriptsize
  \setlength\tabcolsep{4pt}
  \caption{Quantitative results on SynDM; best scores are bolded, second-best scores are in blue.}
  \label{tab:syndmtest}
  \resizebox{\textwidth}{!}{
    \begin{tabular}{l|ccc|ccc|ccc|ccc}
      \toprule
      \textbf{Method} & \multicolumn{3}{c|}{\textbf{Human}} & \multicolumn{3}{c|}{\textbf{Animal}} & \multicolumn{3}{c|}{\textbf{Vehicle}} & \multicolumn{3}{c}{\textbf{Average}} \\
      & FID↓ & PSNR↑ & LPIPS↓ & FID↓ & PSNR↑ & LPIPS↓ & FID↓ & PSNR↑ & LPIPS↓ & FID↓ & PSNR↑ & LPIPS↓ \\
      \midrule
      D3DGS \citep{yang2024deformable}  &\textcolor{blue}{87.83} & 16.97 & \textcolor{blue}{0.305} & 315.3 & 18.55 & \textcolor{blue}{0.272} & 267.8 & 14.70 & 0.543 & \textcolor{blue}{223.6} & 16.74 & 0.373\\
      RoDynRF \citep{liu2023robust}     & 167.3 & 19.66 & 0.318 & \textcolor{blue}{262.0} & 21.00 & 0.302 & 285.3 & 16.31 & \textcolor{blue}{0.395} & 238.2 & 18.99 & \textcolor{blue}{0.338}\\
      DecNeRF \citep{you2024decoupling} & 178.5 & 17.46 & 0.441 & 287.3 & 15.11 & 0.545 & \textcolor{blue}{211.3} & 14.76 & 0.562 & 225.7 & 15.78 & 0.516\\
      DetRF \citep{zhang2025detrf}      & 200.0 & \textcolor{blue}{21.27} & 0.408 & 290.6 & \textcolor{blue}{22.69} & 0.378 & 226.4 & \textcolor{blue}{17.07} & 0.521 & 239.0 & \textcolor{blue}{20.34} & 0.436\\
      \textbf{ExpanDyNeRF (Ours)}       & \textbf{85.61} & \textbf{21.71} & \textbf{0.182} & \textbf{155.8} & \textbf{23.66} & \textbf{0.142} & \textbf{186.7} & \textbf{17.21} & \textbf{0.341} & \textbf{142.7} & \textbf{20.86} & \textbf{0.209}\\
      \bottomrule
    \end{tabular}
  }
  \vspace{-0.15in}
\end{table*}
}

\newcommand{\dynerfQuant}{%
\begin{table}[t]
\centering
\footnotesize
\setlength\tabcolsep{3pt}
\caption{Quantitative results on DyNeRF dataset. We show Coffee and Beef scenes which represent the two primary scenarios in this dataset as shown in the first and second scenes of DyNeRF in Fig.~\ref{fig:comparison}. Best scores are bolded, second-best scores are in blue.}
\label{tab:dynerfQuant}
\begin{tabular}{l|ccc|ccc}
\toprule
\textbf{Method} & \multicolumn{3}{c|}{\textbf{Coffee}} & \multicolumn{3}{c}{\textbf{Beef}} \\
& FID↓ & PSNR↑ & LPIPS↓ & FID↓ & PSNR↑ & LPIPS↓ \\
\midrule
D3DGS \citep{yang2024deformable} & 178.5 & 28.45 & \textcolor{blue}{0.227} & 172.8 & 32.78 & 0.268 \\
RoDynRF \citep{liu2023robust} & 156.3 & 29.13 & 0.268 & \textcolor{blue}{148.2} & 33.45 & \textcolor{blue}{0.252} \\
DecNeRF \citep{you2024decoupling} & \textcolor{blue}{152.8} & 28.87 & 0.304 & 155.4 & 33.12 & 0.285 \\
DetRF \citep{zhang2025detrf} & 165.2 & \textcolor{blue}{29.52} & 0.275 & 151.3 & \textcolor{blue}{33.99} & 0.259 \\
\midrule
\textbf{ExpanDyNeRF (Ours)} & \textbf{132.4} & \textbf{30.32} & \textbf{0.189} & \textbf{135.8} & \textbf{34.92} & \textbf{0.195} \\
\bottomrule
\end{tabular}
\vspace{-0.15in}
\end{table}%
}

\newcommand{\ablationtb}{%
\begin{table*}[t]
\centering
\footnotesize
\setlength{\tabcolsep}{6pt} 
\caption{Quantitative Evaluation of optimization strategies on the SynDM Dataset. Baseline is without any optimization, $L_{nv}^{\sigma}$ only uses density optimization, $L_{nv}^{c}$ only applies color optimization, and $L_{nv}^{\sigma}+L_{nv}^{c}$ is trained with both. $L_{nv}^{\sigma}+L_{nv}^{c}+L_{sr}$ includes all modules including super resolution.}
\label{tab:ablationtb}
\begin{tabular*}{\textwidth}{@{\extracolsep{\fill}}c|ccc|ccc|ccc|ccc}
\toprule
\textbf{Method} & \multicolumn{3}{c|}{\textbf{Human}} & \multicolumn{3}{c|}{\textbf{Animal}} & \multicolumn{3}{c|}{\textbf{Vehicle}} & \multicolumn{3}{c}{\textbf{Average}} \\ 
& FID↓ & PSNR↑ & LPIPS↓ & FID↓ & PSNR↑ & LPIPS↓ & FID↓ & PSNR↑ & LPIPS↓ & FID↓ & PSNR↑ & LPIPS↓  \\ 
\midrule
Baseline  & 200.0 & \textcolor{blue}{21.27} & 0.408 & 290.6 & 22.69 & 0.378 & 226.4 & 17.07 & 0.521 & 239.0 & 20.34 & 0.436\\
$L_{nv}^{\sigma}$ & 158.2 & 21.56 & 0.399 & 237.8 & 23.57 & 0.374 & 186.7 & 16.69 & 0.524 & 194.2 & 20.61 & 0.432\\
$L_{nv}^{c}$ & 147.5 & 20.10 & 0.395 & 162.8 & 19.88 & 0.442 & 210.9 & 16.17 & 0.525 & 173.7 & 18.72 & 0.454\\
$L_{sr}$ & 147.8 & 20.85 & \textcolor{blue}{0.212} & 194.6 & 21.30 & \textcolor{blue}{0.183} & \textcolor{blue}{176.6} & 16.62 & \textcolor{blue}{0.331} & \textcolor{blue}{173.0} & 19.59 & \textcolor{blue}{0.242}\\
\textbf{$L_{nv}^{\sigma}+L_{nv}^{c}$} & \textcolor{blue}{146.2} & \textcolor{blue}{21.66} & 0.395 & \textcolor{blue}{207.9} & \textbf{23.69} & 0.339 & 198.8 & \textcolor{blue}{17.14} & 0.537 & 184.3 & \textcolor{blue}{20.83} & 0.424\\
$L_{nv}^{\sigma}+L_{nv}^{c}+L_{sr}$ & \textbf{85.61} & \textbf{21.71} & \textbf{0.182} & \textbf{155.8} & \textcolor{blue}{23.66} & \textbf{0.142} & \textbf{144.9} & \textbf{17.21} & \textbf{0.304} & \textbf{142.7} & \textbf{20.86} & \textbf{0.209}\\
\bottomrule
\end{tabular*}
\vspace{-0.15in}
\end{table*}%
}

\newcommand{\nvidiatest}{%
\begin{table}[t]
\centering
\footnotesize
\setlength{\tabcolsep}{3pt}
\caption{Quantitative comparison on NVIDIA dynamic scenes. We show Skate and Truck scenes which correspond to the first and second scenes of NVIDIA dataset in Fig.~\ref{fig:comparison}. Best is bold; second-best is blue.}
\label{tab:nvidiatest}
\begin{tabular}{l|ccc|ccc}
\toprule
\textbf{Method} &
\multicolumn{3}{c|}{\textbf{Skate}} &
\multicolumn{3}{c}{\textbf{Truck}} \\
& FID↓ & PSNR↑ & LPIPS↓ & FID↓ & PSNR↑ & LPIPS↓ \\
\midrule
D3DGS \citep{yang2024deformable} & 142.8 & 25.67 & 0.258 & 95.7 & 26.39 & 0.239 \\
RoDynRF \citep{liu2023robust} & 103.5 & 27.89 & 0.087 & 78.4 & 29.13 & 0.063 \\
DecNeRF \citep{you2024decoupling} & 112.9 & 26.83 & 0.134 & 85.6 & 27.56 & 0.115 \\
DetRF \citep{zhang2025detrf} & \textbf{84.34} & \textbf{29.45} & \textbf{0.072} & \textbf{67.69} & \textbf{31.75} & \textbf{0.041} \\
\midrule
\textbf{ExpanDyNeRF (Ours)} & \textcolor{blue}{90.83} & \textcolor{blue}{28.91} & \textcolor{blue}{0.079} & \textcolor{blue}{69.37} & \textcolor{blue}{30.60} & \textcolor{blue}{0.034} \\
\bottomrule
\end{tabular}
\vspace{-0.3in}
\end{table}%
}

\newcommand{\datasetCompare}{%
\begin{table}[t]
\centering
\scriptsize
\setlength\tabcolsep{2pt} 
\caption{ Comparison of dynamic 3D reconstruction datasets. Columns: Multi-view (MV), Deviated View Ground Truth (Deviated View GT), Unconstrained Scene (Unconst. Scene), Camera Motion (Cam. Motion), and Background (Bkg.). SynDM is the only dataset supporting all five attributes.}
\label{tab:datasetCompare}
\begin{tabular}{l|c|c|c|c|c}
\toprule
\textbf{Dataset} & \textbf{MV} & \textbf{Deviated View GT} & \textbf{Unconst. Scene} & \textbf{Cam. Motion} & \textbf{Bkg.} \\
\midrule
DAVIS \citep{pont20172017} & \xmark & \xmark & \cmark & \cmark & \cmark \\ 
iPhone \citep{gao2022monocular} & \xmark & \xmark & \cmark & \cmark & \cmark \\
NeRFDS \citep{yan2023nerf} & \xmark & \xmark & \cmark & \cmark & \cmark \\ 
NVIDIA \citep{yoon2020novel} & \xmark & \xmark & \cmark & \cmark & \cmark \\ 
HyperNeRF \citep{park2021hypernerf} & \xmark & \xmark & \xmark & \cmark & \cmark \\ 
DyNeRF \citep{li2022neural} & \cmark & \cmark & \xmark & \xmark & \cmark \\ 
ActorsHQ \citep{icsik2023humanrf} & \cmark & \cmark & \xmark & \xmark & \xmark \\ 
Multi-face \citep{wuu2022multiface} & \cmark & \cmark & \xmark & \xmark & \xmark \\ 
\midrule
\textbf{SynDM (Ours)} & \cmark & \cmark & \cmark & \cmark & \cmark \\ 
\bottomrule
\end{tabular}
\vspace{-0.1in}
\end{table}%
}

%% file: sec/0_abstract.tex
\begin{abstract}
In dynamic Neural Radiance Fields (NeRF) systems, state-of-the-art novel view synthesis methods often fail under significant viewpoint deviations, producing unstable and unrealistic renderings. To address this, we introduce Expanded Dynamic NeRF (ExpanDyNeRF), a monocular NeRF framework that leverages Gaussian splatting priors and a pseudo-ground-truth generation strategy to enable realistic synthesis under large-angle rotations. ExpanDyNeRF optimizes density and color features to improve scene reconstruction from challenging perspectives. We also present the Synthetic Dynamic Multiview (SynDM) dataset—the first synthetic multiview dataset for dynamic scenes with explicit side-view supervision—created using a custom GTA V-based rendering pipeline. Quantitative and qualitative results on SynDM and real-world datasets demonstrate that ExpanDyNeRF significantly outperforms existing dynamic NeRF methods in rendering fidelity under extreme viewpoint shifts. Further details are provided in the supplementary materials. Code is available at \url{https://github.com/ostadabbas/ExpanDyNeRF}.
\end{abstract}

%% file: sec/1_intro.tex
\section{Introduction}
\label{sec:intro}

Novel view synthesis is essential in applications like mixed reality \citep{xu2023vr,gu2024ue4}, medical supervision \citep{yu2023monohuman,wysocki2024ultra}, autonomous driving \citep{tancik2022block,zhang2024nerf}, and wildlife observation \citep{zhang2023dyn,sinha2023common}. Neural Radiance Fields (NeRF) and their dynamic variants have improved 3D reconstruction with high precision \citep{wang20224k, wang2024hyb, xu2024mega}, speed \citep{garbin2021fastnerf,gao2024general,lee2024sharp}, and style editing \citep{gu2021stylenerf,chen2024upst}. Alternatively, Gaussian splatting \citep{4dgs,wu2024recent, yang2024deformable} offers an efficient and flexible framework for high-quality rendering. While both approaches produce sharp results from primary viewpoints, renderings degrade with significant viewpoint shifts due to the lack of diverse-view supervision during training, which is a limitation of monocular settings (Fig.~\ref{fig:comparison}).

\comparison

To overcome these limitations, we propose Expanded Dynamic NeRF (ExpanDyNeRF), a dynamic NeRF framework designed to expand reliable rendering to large-angle novel views, even under monocular camera constraints. This end-to-end pipeline, illustrated in Fig.~\ref{fig:frame}, incorporates a novel-view pseudo ground truth strategy that optimizes model training from novel view by leveraging Gaussian priors \citep{wang2024freesplat}, effectively refining dynamic object contours and color consistency across frames.

One of the key challenges in novel view synthesis for dynamic scenes lies in the lack of suitable datasets that offer both dynamic motion and side-view supervision. Existing datasets either have dynamic camera motion without side-view ground truth (e.g., NVIDIA \citep{yoon2020novel}) or include rotated views without camera motion (e.g., DyNeRF \citep{li2022neural}). This gap stems from the difficulty of capturing multi-view dynamic scenes in the real world. To address this, we introduce SynDM, a GTA V-based dataset with a novel dynamic camera dome that enables synchronized main-view motion and side-view supervision. Our evaluation focuses on SynDM, with only qualitative comparisons on existing datasets. Our main contributions are:

\begin{itemize}
    \item We identify and characterize the limitations of current monocular dynamic NeRFs in rendering from significantly deviated viewpoints, highlighting their inability to preserve structure and appearance consistency under angular shifts
    \item We propose \textbf{ExpanDyNeRF}, a novel dynamic NeRF architecture that incorporates pseudo-novel view supervision using Gaussian splatting priors, enabling reliable synthesis at large viewpoint deviations
    \item We introduce \textbf{SynDM}, the first synthetic dataset for dynamic monocular NeRFs with paired primary and rotated views, captured via a custom GTA V pipeline to benchmark novel view synthesis under controlled angular perturbations 
    \item We perform comprehensive experiments on the SynDM, DyNeRF, and NVIDIA datasets, demonstrating improved perceptual quality and geometric consistency over previous dynamic NeRF methods, especially when handling large viewpoint deviations
\end{itemize}

%% file: sec/2_relatedwork.tex
\section{Related Work}
\label{sec:relatedwork}
\textbf{NeRF-based Dynamic Novel View Synthesis.} 
NeRF algorithms have emerged as a powerful technique for high-quality 3D scene reconstruction from a sparse set of images. Original NeRF \citep{mildenhall2021nerf} leverages a fully connected deep neural network to model the volumetric scene function. This function outputs the color and density for any given 3D point and viewing direction, enabling the synthesis of novel views through volume rendering techniques. NeRFs have been particularly successful in static scenes, and recent advancements have extended their application to dynamic 3D reconstruction. Dynamic NeRF methods, such as HyperNeRF \citep{park2021hypernerf}, DetRF \citep{zhang2025detrf}, and DecNeRF \citep{you2024decoupling} incorporate temporal components, allowing for the modeling of scenes with moving objects and varying illumination. These methods often employ additional strategies like temporal consistency loss and motion field modeling to handle the complexities of dynamic environments. Despite their success, dynamic NeRFs face challenges with computational expense and the need for densely sampled temporal data. Many dynamic NeRF methods require substantial compute resources—e.g., Dynibar \citep{li2023dynibar} reports training times exceeding 1 day on 8 A100 GPUs for a single scene—highlighting the difficulty of scaling dynamic NeRFs to large datasets or real-time applications.
\framework

\textbf{Gaussian Splatting-based Dynamic Novel View Synthesis.}
3D Gaussian splatting \citep{kerbl20233d} offers an efficient alternative to NeRF by representing scenes with Gaussian blobs, enabling real-time, high-resolution rendering. Recent methods like 4DGS \citep{4dgs} and D3DGS \citep{yang2024deformable} extend this approach to dynamic scenes using monocular inputs, capturing non-rigid motion via deformation fields. While NeRF provides high-fidelity reconstructions and Gaussian splatting excels in speed, both struggle with novel view synthesis from deviated angles in monocular settings. Our method combines their strengths to overcome these limitations.

%% file: sec/3_method.tex
\section{Method}
\label{method}
We present ExpanDyNeRF, a monocular dynamic NeRF framework for synthesizing novel views of 3D scenes under large viewpoint deviations. To address the lack of ground truth in monocular settings, we combine a two-branch dynamic NeRF with pseudo-supervision from 3D Gaussian priors. Section~\ref{dynerf} details the dynamic NeRF backbone, Section~\ref{density} describes pseudo-novel view supervision using 3D Gaussian priors, and Section~\ref{syndm} introduces our SynDM dataset with multi-view dynamic scenes.

\subsection{ExpanDyNeRF Model Architecture}
\label{dynerf}

Our preliminary experiments indicate that NeRF provides greater visual consistency than Gaussian Splatting, especially for distant elements like the sky (shown in Fig.~\ref{fig:ablationGSNeRF}). Therefore, we adopt NeRF as the backbone of our model.

\ablationGSNeRF

\textbf{NeRF vs Gaussian Splatting Analysis:} While both methods appear visually consistent in primary views, significant differences emerge in side view reconstruction. As demonstrated in Fig.~\ref{fig:ablationGSNeRF}, 3DGS introduces substantial artifacts in side views, with the sky being incorrectly reconstructed as nearby structures, obstructing distant background elements including mountains and bushes. In contrast, Instant-NGP (NeRF-based) retains the expected characteristics of the sky as distant and uniform, achieving higher fidelity and richer scene details. Accordingly, we adopt a NeRF backbone for its stability under large viewpoint rotations, while using Gaussian Splatting only as a prior generator rather than a rendering backbone. We do not claim NeRF to be universally superior, but find this combination effective for the targeted setting.

Following the architecture in \citep{zhang2025detrf}, we employ two intertwined neural networks: 
\(\Phi_{b}\)  for modeling the static background and \(\Phi_{f}\) for the dynamic foreground. As illustrated in Fig.~\ref{fig:frame}, our framework is structured into two main components: (1) a backbone dynamic NeRF model that processes rays to extract density (\(\sigma\)) and color (\(c\)) features from both background and foreground models, generating rendering predictions \(\hat{I}_{t}\) from primary camera positions, supervised by the super-resolution loss \(\mathcal{L}_{sr}\); (2) novel view feature optimization that uses Gaussian splatting priors to generate 3D representations for each frame, facilitating optimization of density and color features via pseudo ground truth for novel views. This includes updating \(\sigma_{f}\) and \(c_{f}\) using novel view loss metrics \(\mathcal{L}_{nv}^{\sigma}\) and \(\mathcal{L}_{nv}^{c}\), respectively, enhancing feature representation across different perspectives.

\textbf{Preliminaries:} We use $N$ video frames $I_{t}, t \in [1, N]$, to reconstruct a point cloud and estimate primary camera poses $P$. For each pose $P_t \in P$, we compute ray trajectories to sample points $\mathbf{x} = (x, y, z)$ along ray $r$ at time $t$ in direction $\mathbf{d}$. The sampling process is defined as $\mathcal{F}(P_t) \rightarrow (\mathbf{x},\mathbf{d})$, linking the camera orientation to the sampled points.

\textbf{Static Background Representation:}
The static background module \(\Phi_{b}\) takes all $N$ frames and encodes static scene components using a distribution-based representation. This improves alignment between camera projections and the background geometry, allowing simplified renderings of low-variance static features. The module, \( \Phi_{b}(\mathcal{F}(P), \theta) \to (c_{b}, \sigma_{b})\), predicts color $c_{b}$ and density $\sigma_{b}$ of spatial points from all poses in $P$, using a prior distribution (\( \theta \sim P_{\Theta}(\theta) \)).


\textbf{Dynamic Foreground Representation:}
The dynamic foreground module \(\Phi_{f}\) captures temporal variation using a three-frame sliding window. It integrates spatial coordinates, viewing directions, and the timestamp $t$: (\(\Phi_{f}(\mathcal{F}(P_t), t) \rightarrow (c_{f}, \sigma_{f})\)). Temporal consistency is maintained by encoding time \( t \) directly and using optical flow to estimate scene flow, predicting future states of dynamic objects. Continuity constraints are applied to maintain smooth attribute transitions across frames, expressed as: \(\mathcal{L}_{cont} = \sum \| \sigma_{f}(t+1) - \sigma_{f}(t) \|^{2}\).


\textbf{Primary View Reconstruction Loss:} 
The system employs a reconstruction loss to optimize \(\Phi_{b}\) and \(\Phi_{f}\) by minimizing the discrepancies between the features $\hat{C}(r)$ from rendered images and $C(r)$ from the ground truth images, defined as $\mathcal{L}_{rec} = \sum_{i=1}^{N} \sum_{r \in R} \| \hat{C}(r) - C(r) \|_2^2$. This loss ensures the renderings from the primary views closely match the ground truth frames, setting up a baseline for the following novel view optimization.

\textbf{Super-Resolution Loss:}
Inspired by SOTA super-resolution methods \citep{realesrgan, nerfsr}, we incorporate a super-resolution loss ($\mathcal{L}_{sr}$) to enhance image quality. 
Rendered patches from ExpanDyNeRF are processed by a pre-trained super-resolution model, which preserves fine textures while increasing resolution. The loss is computed by comparing sampled patches from the predicted and corresponding high-resolution reference. The formulation of $\mathcal{L}_{sr}$ is:

{\footnotesize
\[
\mathcal{L}_{sr} = \sum_{k=1}^{K} \left\| \hat{Q}_k - Q_k \right\|_1 
+ \sum_{k=1}^{K}\sum_{l}\lambda_{l} \left\| F_{vgg}^{l}(\hat{Q}_k) 
- F_{vgg}^{l}(Q_k) \right\|_1.
\]
}

Here, \( \hat{Q}_k \) and \( Q_k \) represent the super-resolution prediction and reference patches, respectively, $F_{vgg}^{l}$ is a set of layers in a pretrained VGG-19 feature extractor, and $\lambda_l$ is the reciprocal of the number of neurons in layer $l$, combining reconstruction and perceptual losses.

\novelView

\subsection{Pseudo Ground Truth Optimization Strategy}
\label{density}
Through empirical experiments, we observed that foreground objects appear blurrier than the background during viewpoint rotation. This is due to affine distortion, where nearby objects exhibit greater apparent motion than distant ones. To address this, we prioritize optimizing foreground representations. Our method leverages FreeSplatter \citep{wang2024freesplat} to generate high-quality 3D mesh priors, enabling pseudo-ground truth supervision from novel viewpoints.

\textbf{Pseudo Ground Truth Generation for Novel Views}:
For each input frame $I_t$, we construct a 3D Gaussian prior representing the foreground object in its local coordinate frame. Centered around this object, we define a dome-shaped sampling space, with a radius \(R_{d}\) as shown in Fig.~\ref{fig:novelView}. The radius \(R_{d}\) represents the distance from the primary viewpoint to the object. The position of $P_t$ on the dome is denoted as \( (elevation = e, azimuth = 0, radius = R_{d}) \), where $e$ corresponds to the elevation angle of the primary recording view. \newline
\textbf{Novel Viewpoint Sampling Strategy:} 
We systematically sample novel viewpoints by varying both azimuth and elevation angles while maintaining the fixed radius \(R_{d}\). Specifically, we generate viewpoints spanning azimuth angles from \( -45^{\circ} \) to \( 45^{\circ} \) in \( 5^{\circ} \) increments, at three elevation levels: \( 0^{\circ} \), \( 15^{\circ} \), and \( 30^{\circ} \). The forward vector of all camera poses \( P_{nv}^{(d)} \) points towards the dome center, ensuring consistent object framing across viewpoints. From these novel camera poses, we render pseudo ground truth images that encode the expected density and color distributions of the foreground object. The resulting renderings provide both RGB color and corresponding shape masks at different viewing angles, creating comprehensive supervision that would be impossible to obtain from real monocular capture. \newline
\textbf{Mapping Novel Views to the NeRF Coordinate System}: 
To apply pseudo ground-truth supervision at novel viewpoints, we transform the sampled camera poses from the Gaussian prior coordinate system to the NeRF coordinate system using a rigid alignment matrix. Let the primary camera pose $P_t$ in the foreground NeRF coordinate system be $P_t^{(n)}$, and its corresponding pose in Gaussian prior coordinate system be $P_t^{(d)}$. The transformation matrix \( T \) that aligns the two coordinate systems is computed as: $T = P_t^{(n)} \cdot (P_t^{(d)})^{-1}$. We apply this transformation,  \( T \), to each novel view camera pose,  \( P_{nv}^{(d)} \), sampled in the Gaussian prior coordinate system, to transfer all new camera positions to the foreground NeRF coordinate system:$P_{nv} = \{P \cdot T, \forall P \in P_{nv}^{(d)}\}.$ \newline
\noindent\textbf{Novel View Loss}: 
During each training iteration, two symmetric novel views are randomly sampled per frame from \( P_{nv} \). For each selected novel viewpoint, a set of rays $\mathbf{R}_{nv}$ is sampled from the camera pose. Color and density predictions in the foreground NeRF are obtained by evaluating the network $\Phi_{f}$ on sampled ray $(\mathcal{F}(P_{nv}), t)$, producing outputs $(c_{f}, \sigma_{f})$. Specifically, $\mathcal{F}(P_{nv}) \rightarrow (\mathbf{x},\mathbf{d})$ samples points along a ray $r_{nv} \in \mathbf{R}_{nv}$. We then integrate the predicted color and density values along each ray $r_{nv}$ to produce the corresponding pixel-wise predictions $\hat{C}_{f}(r_{nv})$ and $\hat{\sigma}_{f}(r_{nv})$. The corresponding novel view loss is defined as:
\begin{equation}
\begin{aligned}
\mathcal{L}_{nv} &= \mathcal{L}_{nv}^{c} + \mathcal{L}_{nv}^{\sigma} \\
&= \sum_{r \in R_{nv}} \big(
\| \hat{C}(r_{nv}) - C(r_{nv}) \|_2^2 +
\| \hat{\sigma}(r_{nv}) - \sigma(r_{nv}) \|_2^2
\big). \nonumber 
\end{aligned}
\end{equation}


where $C(r_{nv})$ and $\sigma(r_{nv})$ denote the pseudo ground truth color and density values for ray $r_{nv}$, respectively. This loss encourages the model to match the rendered appearance and structure of the pseudo-supervised novel views, improving generalization to unseen angles. A further explanation and visualization of ray sampling strategies are detailed in Fig. \ref{fig:padding} in the supplementary. To manage the risk of exploding gradients early in training, we defer inclusion of $\mathcal{L}_{nv}$ until after a fixed number of epochs. 
The final total loss function is given by:
\[
   \mathcal{L} = \mathcal{L}_{cont} + \mathcal{L}_{rec} + \mathcal{L}_{sr} + \mathcal{L}_{nv}.
\]

\subsection{Synthetic Dynamic Multiview (SynDM)  Dataset }
\label{syndm}

To enable quantitative evaluation of novel view synthesis under significant viewpoint deviations, we introduce our SynDM dataset, built using the high-fidelity simulation platform GTA V. The game offers rich dynamic environments and realistic rendering, making it an ideal foundation. However, a core limitation of GTA V is its support for only a single active viewport, which posing a challenge for synchronized multi-view dynamic scene capture. We extend the GTAV-TeFS \citep{luo2023temporal} framework--originally developed for dual-camera capture --into a generalized multi-camera pipeline to simultaneously support both monocular primary camera capture and multi-view stereo camera collection in GTA V's dynamic environment. Traditionally, collecting data from multiple camera views in a single-viewport engine requires frame swapping, where  each camera is rendered sequentially. Under a 60 Hz refresh rate, this results in a latency of at least 16.7 ms per camera swap. While acceptable for static scenes, this approach quickly breaks down in dynamic settings and as we add cameras, as the accumulated latency introduces motion misalignment and temporal artifacts. To address this we developed a custom plugin that semi-freezes the game's graphical state while allowing the rendering and physics engine to continue running. This design enables us to cycle through camera views in a controlled and consistent manner during a single logical frame. With precise scheduling, we reduced the per-swap latency from 16.7ms to just 0.2ms, making high-resolution, low-latency multi-view capture of dynamic scenes possible.

Our dataset enables synthetic object tracking via synchronized multi-view recordings, offering a robust ground truth for evaluating dynamic NeRF-based novel view synthesis. It consists of nine distinct scenes spanning three categories--humans, vehicles, and animals (Fig.~\ref{fig:gallery}). Each scene is captured using 22 cameras: 19 are distributed horizontally around a reference point at $5^\circ$ intervals from $-45^\circ$ to $45^\circ$, including a central anchor camera, while the remaining three are elevated vertically at $-45^\circ$, $0^\circ$, and $45^\circ$. All frames are rendered at a resolution of 1920×1080 with a $90^\circ$ horizontal and $59^\circ$ vertical field of view.

\datasetCompare
\textbf{Necessity and Advantage of Proposed SynDM Dataset}  As shown in Table~\ref{tab:datasetCompare}, existing datasets lack critical features for evaluating dynamic novel view synthesis under large viewpoint deviations. Most importantly, no real-world dataset provides deviated view ground truth, severely limiting quantitative assessment beyond primary viewpoints. Our SynDM dataset uniquely combines all essential features, including multi-view data, deviated GT, full-scene representation, and camera motion, enabling comprehensive evaluations previously impossible with existing datasets alone.

\SynDM

%% file: sec/4_results.tex
\section{Experimental Results}
\label{experiment}
We present a comprehensive evaluation of ExpanDyNeRF against state-of-the-art dynamic novel view synthesis methods. Our evaluation strategy progresses from controlled synthetic environments to challenging real-world scenarios, demonstrating robustness across diverse settings. More results can be find in our Supplementary Material. 
\GTA
\syndmtest
\subsection{Experimental Setup}
\label{experimentalsetup}

\textbf{Datasets.} We evaluate ExpanDyNeRF on three datasets with complementary characteristics. Our \textbf{SynDM Dataset} provides complete ground truth for quantitative analysis across three scene categories (human, animals, vehicles) spanning rural and urban environments within GTA V simulation. For training, we use the first 24 frames from each scene and evaluate on 12 novel views, uniformly sampled between $-30^\circ$ to $+30^\circ$ at $5^\circ$ intervals. The \textbf{DyNeRF Dataset} \cite{li2022neural} offers real-world dynamic scenes with multi-view ground truth, enabling quantitative evaluation under challenging viewing conditions. The \textbf{NVIDIA Dataset} \citep{yoon2020novel} provides real-world monocular sequences for generalization assessment, though without deviated viewpoint ground truth (see Table \ref{tab:datasetCompare} for detailed comparison).

\textbf{Implementation Details.}
\textbf{Implementation Details.}
ExpanDyNeRF is trained with per-scene optimization on 2 A100 GPUs for 300k iterations (approximately 15 hours per scene). The training cost is primarily dominated by NeRF optimization with pseudo-novel view supervision, while Gaussian prior generation is performed once per frame as a preprocessing step and incurs marginal overhead. Consistent with prior dynamic NeRF approaches, our method prioritizes rendering fidelity under large viewpoint deviations over training efficiency. Loss coefficients are set to: $\lambda_{nv}^{c} = 1.0$, $\lambda_{nv}^{\sigma} = 0.1$, and $\lambda_{sr} = 0.5$. All other parameters follow \citep{d4nerf}.


\textbf{PSNR Limitation Analysis.} Our experiments reveal important limitations of PSNR in evaluating perceptual quality. Despite ExpanDyNeRF producing sharper, more detailed reconstructions, PSNR scores may paradoxically be lower due to localized high-density errors highlighted in pixel-wise error heatmaps. In contrast, baseline methods with blurry results yield smoother transitions, leading to lower MSE despite poorer visual quality. This occurs because blurry regions (e.g., object boundaries, fine textures) blend into backgrounds, minimizing MSE contributions (shown in Fig.~\ref{fig:heatmap} in the supplementary). This phenomenon demonstrates why complementary perceptual metrics like LPIPS and FID are essential for comprehensive quality assessment, particularly when evaluating sharpness and clarity improvements.

\subsection{Evaluation on SynDM Dataset}
\label{testsyndm}

The SynDM dataset enables comprehensive quantitative evaluation with complete ground truth across diverse dynamic scenes. We compare ExpanDyNeRF against four state-of-the-art methods using PSNR, LPIPS, and FID metrics (Table~\ref{tab:syndmtest}).

\textbf{Quantitative Results.} ExpanDyNeRF achieves the highest PSNR score of 20.86 while producing the sharpest renderings. Although DetRF also reports competitive PSNR despite producing heavily blurred images, this underscores the importance of complementary perceptual metrics. Our model significantly outperforms competing methods in both LPIPS (38\% lower than second-best) and FID (36\% lower than second-best), demonstrating superior perceptual alignment and distributional consistency.

\textbf{Qualitative Analysis.} Fig.~\ref{fig:GTA} illustrates ExpanDyNeRF's superior shape coherence and color stability in dynamic regions. Each baseline method exhibits distinct failure patterns: DecNeRF renders sharp details but suffers from poor depth perception, resulting in flat, cardboard-like appearances that fail to maintain 3D structure consistency. D3DGS introduces depth inconsistencies causing foreground objects to fracture under rotation, particularly evident after $30^\circ$ viewpoint changes where object parts appear disconnected. RoDynRF struggles with consistent object placement, often producing floating artifacts or misaligned body parts. These failure modes stem from insufficient side-view supervision during training. Further visualizations of results are provided in Fig. \ref{fig:suppchicken} and \ref{fig:suppmale} in the supplementary.

\subsection{Evaluation on DyNeRF Dataset}
\label{testdynerf}
\dynerfQuant
\nvidiatest

The DyNeRF dataset provides real-world dynamic scenes with multi-view ground truth, but presents challenges for monocular training due to its stationary camera setup. Unlike our SynDM dataset which provides natural camera motion, DyNeRF's fixed camera positions prevent direct COLMAP reconstruction from monocular sequences. To address this limitation, we construct synthetic monocular sequences by selecting frames from central cameras (cam0, cam4, cam5, cam6) across different timestamps, creating the necessary camera motion for COLMAP initialization. We then adopt a holdout validation strategy using these central cameras for training while reserving the geometrically challenging outer cameras (cam01 and cam10) as test views—specifically selected as the most deviated viewpoints from the training set's central viewing positions.

\textbf{Quantitative Results.} Table~\ref{tab:dynerfQuant} shows our method's consistent superiority across all baseline methods on representative DyNeRF scenes. ExpanDyNeRF achieves the best performance across all metrics in both scenes. Our method demonstrates particularly strong performance on Coffee (FID: 132.4, PSNR: 30.32, LPIPS: 0.189) and Beef sequences (FID: 135.8, PSNR: 34.92, LPIPS: 0.195), with consistent improvements across all three evaluation metrics in both representative scenarios.

\textbf{Qualitative Analysis.} As shown in Fig.~\ref{fig:comparison}, our method maintains superior performance on DyNeRF scenes. Leading dynamic NeRF and Gaussian splatting methods—DetRF, RoDynRF, DecNeRF, and D3DGS—all suffer from artifacts and depth errors at viewpoints distant from the training pose, whereas ExpanDyNeRF maintains accurate shape and color consistency. The improvements are especially significant for the challenging viewpoints, which demand accurate geometric reasoning due to their significant displacement from training views, confirming that our pseudo-ground truth supervision strategy successfully addresses fundamental geometric consistency challenges in dynamic 4D scene reconstruction.

\subsection{Generalization to NVIDIA Dataset}
\label{testnvidia}

The NVIDIA dataset provides real-world monocular sequences captured with stationary cameras, offering insights into our method's generalization capabilities under different capture conditions. However, this dataset presents limitations for evaluating large viewpoint deviations: the 12 cameras are positioned in a compact matrix formation with relatively small angular separations, lacking the significant viewpoint variations needed for rigorous novel view synthesis evaluation. Consequently, unlike DyNeRF and SynDM datasets, NVIDIA lacks ground truth for challenging side-view positions, limiting quantitative assessment to modest viewpoint changes. As a result, performance on the NVIDIA dataset should be interpreted as a complementary indicator of real-world robustness, rather than a definitive evaluation of large-angle novel view synthesis.

\ablationNVLossNew
\ablationtb

\textbf{Quantitative Analysis.} Table~\ref{tab:nvidiatest} shows competitive performance on representative NVIDIA scenes. While we do not achieve the highest scores, this reflects the dataset's unique characteristics: 12 stationary cameras (using for both train and test) positioned primarily in front of the scene, favoring methods optimized for primary viewpoints. Our approach demonstrates strong performance on Skate (FID: 90.83, PSNR: 28.91, LPIPS: 0.079), and Truck (FID: 69.37, PSNR: 30.60, LPIPS: 0.034), consistently achieving second-best results. DetRF achieves superior performance on primary viewpoints due to its focus on depth estimation from forward-facing cameras, while our method's relatively balanced performance across different viewpoints suggests more stable behavior beyond primary training views. This trade-off validates our design choice to prioritize robustness over dataset-specific optimization.

\textbf{Qualitative Results.} Fig.~\ref{fig:comparison} reveals notable limitations in existing methods following camera rotation. We compare against leading dynamic NeRF and Gaussian splatting methods including DetRF \cite{zhang2025detrf}, RoDynRF \cite{sabour2023robustnerf}, DecNeRF \cite{you2024decoupling}, and D3DGS \cite{yang2024deformable}. All baseline methods suffer from artifacts and depth errors at viewpoints distant from training poses. Specifically, RoDynRF struggles to maintain structural consistency—rendered figures may retain correct foot placement yet exhibit misaligned upper bodies that appear "stuck" to background elements like pillars in the skate scene. DecNeRF exhibits cardboard-like foreground appearance or complete disappearance in side views, reflecting insufficient depth estimation that leads to extremely thin or missing foreground rendering from oblique angles. D3DGS shows spatial fragmentation stemming from absent side-view supervision, hindering accurate depth estimation. In contrast, ExpanDyNeRF maintains accurate shape and color consistency across the evaluated rotation angles, effectively addressing common failure modes such as fractured object reconstructions and temporal inconsistency. Further visualizations of results on the NVIDIA dataset are provided in Figs. \ref{fig:suppskate} and \ref{fig:supptruck} in the supplementary.

\subsection{Ablation Study}
\label{ablation1}

We conduct a comprehensive ablation study to validate the contribution of each component in our optimization strategy. The analysis examines the effects of our novel view losses and super-resolution enhancement.

\textbf{Component Analysis.} Fig.~\ref{fig:ablationNVLoss} demonstrates the visual impact of different loss combinations on novel view synthesis quality. The baseline model without novel view supervision produces poorly defined shapes and noticeable blurring, indicating the critical importance of side-view guidance. Introducing only color-based novel view loss $L_{nv}^{c}$ reduces blurring but introduces white artifacts due to absent shape constraints provided by the density supervision. Conversely, using only $L_{nv}^{\sigma}$ helps preserve object shape but results in faded or distorted colors. The synergistic effect between $L_{nv}^{\sigma}$ and $L_{nv}^{c}$ is evident when combined, leading to improved structural clarity and temporal stability, although fine textures remain underrepresented. The super-resolution loss $L_{sr}$ proves essential for texture enhancement but requires the foundation of geometric consistency from novel view losses. Using $L_{sr}$ in isolation produces artifacts and inconsistencies, confirming that high-level semantic supervision must precede low-level texture refinement. Importantly, $L_{sr}$ primarily improves perceptual sharpness and does not introduce new geometric constraints. Without the geometric consistency established by $L_{nv}^{\sigma}$ and $L_{nv}^{c}$, it tends to amplify existing structural errors rather than correct them.

\textbf{Quantitative Validation.} Table~\ref{tab:ablationtb} confirms these visual observations across all metrics. The best performance is achieved when integrating $L_{nv}^{\sigma}$, $L_{nv}^{c}$, and super-resolution loss $L_{sr}$. Notably, using $L_{sr}$ in isolation fails to produce acceptable renderings, underscoring the essential role of novel view supervision in mitigating artifacts during viewpoint rotations. The full combination achieves optimal perceptual quality, confirming our optimization strategy's effectiveness in reducing visual artifacts and enhancing overall realism.

%% file: sec/5_conclusion.tex
\section{Discussion and Conclusion}
\textbf{Limitations.} 
Despite outperforming existing models, ExpanDyNeRF has limitations, particularly in handling extreme viewing angles (beyond 45 degrees) and unseen background generation. Also, ExpanDyNeRF requires per-scene optimization, resulting in higher computational cost compared to feed-forward or purely Gaussian-based methods.

\noindent \textbf{Conclusion.} 
ExpanDyNeRF advances dynamic NeRF by significantly improving novel view synthesis, particularly at wider viewing angles, by extending the range of stable visualization. Our SynDM dataset, based on GTA V for dynamic multiview scenarios, provides a strong foundation for evaluating dynamic scene reconstructions from varied angles. Our evaluations demonstrate ExpanDyNeRF's superior ability to render dynamic scenes.

%% file: sec/X_suppl.tex
\clearpage
\setcounter{page}{1}
\maketitlesupplementary
\textbf{Overview of Supplementary Materials}
This supplementary document provides extended experimental results, ablations, and visual comparisons that complement the main paper. In particular, we include detailed ablation studies on our ray sampling and padding strategies (Fig. \ref{fig:padding}) and error heatmaps illustrating the limitations of standard metrics such as PSNR and MSE (Fig. \ref{fig:heatmap}). Furthermore, we include additional novel view synthesis visualizations across both the SynDM and NVIDIA datasets in Figs. \ref{fig:suppchicken}--\ref{fig:supptruck}. These results further validate the effectiveness of ExpanDyNeRF in capturing sharper, more consistent scene details across diverse scenarios.  For an interactive overview with richer visualizations, readers are encouraged to view the accompanying \texttt{IO} page, which showcases additional qualitative examples and video results.
\newline
\newline
\padding
\textbf{Ray Sampling Strategies}
We compared various ray sampling strategies for novel view density and color optimization in Equation~\ref{density}. Examples are shown in Fig.~\ref{fig:padding}.
Global sampling over the whole frame yields results in Panel $(e)$ similar to the base output in Panel $(a)$, due to the small proportion of dynamic segments in the frame, causing generalized and ineffective updates.
Alternate strategies sample within the foreground object's area shown white in Panel $(c)$, which may overlook updates outside this zone. Panel $(f)$ demonstrates that sampling from various viewpoints for dynamic density updates can unintentionally extend beyond the intended mask, causing non-dynamic areas to obscure the background.
Panel $(g)$ shows the third strategy where the GaussianBlur \citep{gonzalez2009digital} expands the foreground boundary, creating a zero gray-scale edge. Sampling within this blurred mask improves results, yet areas adjacent to the person still see undue dynamic density updates beyond the motion mask.
Our final strategies focused on ray sampling within the padded area of the mask's bounding box (bounding boxes in Panel $(c)$), which outperforms the other strategies. Experimentation showed that while larger padding, like 10 pixels in Panel $(h)$, achieves comparable foreground optimization to smaller padding, such as 2 pixels in Panel $(d)$, it adversely affects background clarity.

\heatmap

\suppTwo
\suppThree
\suppFour
\suppFive